\documentclass{article}
\usepackage{ijcai18}

\usepackage{times}
\usepackage{soul}
\usepackage{url}
\usepackage[utf8]{inputenc}
\usepackage[small]{caption}
\usepackage{graphicx}
\usepackage{amsmath}
\usepackage{amssymb}
\usepackage{multirow}
\usepackage{amsmath}
\usepackage{url}
\title{Anonymizing $k$-Facial Attributes via Adversarial Perturbations}

\author{
Saheb Chhabra$^1$, 
Richa Singh$^1$, 
Mayank Vatsa$^1$ \textnormal{and}
Gaurav Gupta$^2$
\\ 
$^1$ IIIT Delhi, New Delhi, India \\
$^2$ Ministry of Electronics and Information Technology, New Delhi, India\\
\{sahebc, rsingh, mayank@iiitd.ac.in\}, gauravg@gov.in
}
\begin{document}

\maketitle

\begin{abstract}
A face image not only provides details about the identity of a subject but also reveals several attributes such as gender, race, sexual orientation, and age. Advancements in machine learning algorithms and popularity of sharing images on the World Wide Web, including social media websites, have increased the scope of data analytics and information profiling from photo collections. This poses a serious privacy threat for individuals who do not want to be profiled. This research presents a novel algorithm for anonymizing selective attributes which an individual does not want to share without affecting the visual quality of images. Using the proposed algorithm, a user can select single or multiple attributes to be surpassed while preserving identity information and visual content. The proposed adversarial perturbation based algorithm embeds imperceptible noise in an image such that attribute prediction algorithm for the selected attribute yields incorrect classification result, thereby preserving the information according to user's choice. Experiments on three popular databases i.e. MUCT, LFWcrop, and CelebA show that the proposed algorithm not only anonymizes \textit{k}-attributes, but also preserves image quality and identity information. 
\end{abstract}

\section{Introduction}
\textit{``The face is the mirror of the mind, and eyes without speaking confess the secrets of the heart." - Letter 54, St. Jerome}. 

\begin{figure}
\centering
    \includegraphics[scale=0.5]{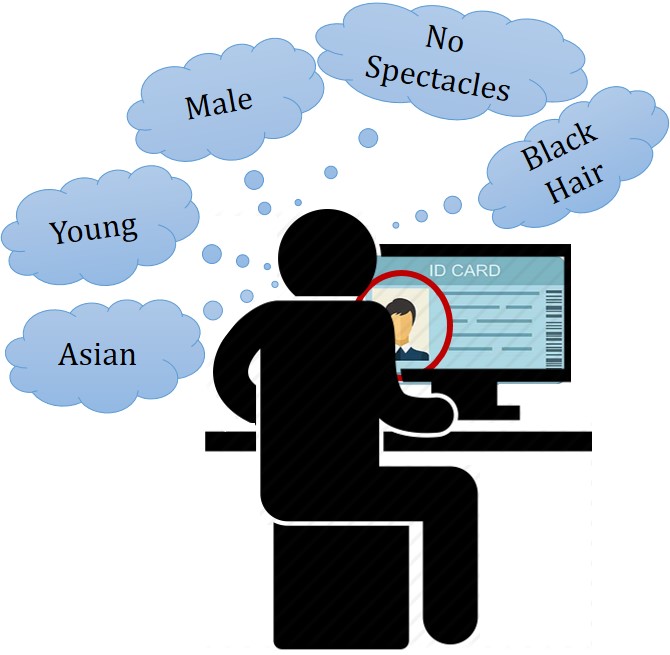}
	\caption{Profiling of a person using his face image in ID card for malicious purpose.}
\label{fig:Introduction_Image}
\end{figure}

Face analysis has been an area of interest for several decades in which researchers have attempted to answer important questions ranging from identity prediction \cite{zhou2017multiple}, to emotion recognition \cite{li2016analysis}, and attribute prediction \cite{abdulnabi2015multi}, \cite{sethi2018residual}. Focused research efforts and use of deep learning models have led to a high performance for tasks involved in face analysis. For instance, state-of-the-art results on YouTube Faces \cite{wolf2011face} and Point and Shoot Challenge \cite{beveridge2013challenge} databases are over 95\%, and 97\% \cite{goswami2017face}, and attribute recognition on the Celeb-A database is more than 90\% \cite{sethi2018residual}. While there are several advantages of these high accuracies, they also pose a threat to the privacy of individuals. As shown in Figure \ref{fig:Introduction_Image}, several facial attributes such as age, gender, and race can be predicted from one's profile or social media images. In a recent research, Wang and Kosinksi predicted the ``secrets of the heart'', such as predicting \textit{sexual orientation} from face images \cite{wang2017deep}. They reported 81\% accuracy for differentiating between gay and heterosexual men with a single image and 74\% accuracy is achieved for women. Similarly, targeted advertisements by predicting the gender and age from the profile and social media photographs have been a topic of research for the last few years. 

%%%%%%%%%% Introduction Image %%%%%%%%%%%%%%%%%%

%%%%%%%%%%%%%%%%%%%%%%%%%%%%%%%%%%%%%%%%%%%%%%%%

%%%%%%%%%%%%%%%%%%% Related Work Table %%%%%%%%%%%%%%%%%%%%
\begin{table*}[]
\small
\centering

\begin{tabular}{|c|c|c|c|c|c|}
\hline
\textbf{Author}    & \textbf{Method}                                                   & \textbf{\begin{tabular}[c]{@{}c@{}}No. of Attributes\end{tabular}} & \textbf{Dataset}                                            & \textbf{\begin{tabular}[c]{@{}c@{}}Controlling \\ Attributes\end{tabular}} & \textbf{\begin{tabular}[c]{@{}c@{}}Visual \\ Appearance\end{tabular}} \\ \hline
Othman and Ross, 2014                                                                                      & \begin{tabular}[c]{@{}c@{}}Face Morphing \\ and fusion\end{tabular}               & One                        & MUCT                                                                   & No                              & Partially Preserved        \\ \hline
Mirjalili and Ross, 2017                                                                                          & \begin{tabular}[c]{@{}c@{}}Delaunay Triangulation \\ and fusion\end{tabular}      & One                        & MUCT, LFW                                                              & No                              & Partially Preserved        \\ \hline
Mirjalili \textit{et al.}, 2017                                                          & \begin{tabular}[c]{@{}c@{}}Fusion using \\ Convolutional Autoencoder\end{tabular} & One                        & \begin{tabular}[c]{@{}c@{}}MUCT, LFW, \\ Celeb-A, AR-Face\end{tabular} & No                              & Preserved                  \\ \hline
\begin{tabular}[c]{@{}c@{}}Rozsa \textit{et al.}, 2016, 2017\end{tabular} & Fast Flipping Attribute                                                           & Multiple                   & CelebA                                                                 & No                              & Preserved                  \\ \hline
Proposed                                                                                                                       & \begin{tabular}[c]{@{}c@{}}Adversarial \\ Perturbations\end{tabular}              & Multiple                   & \begin{tabular}[c]{@{}c@{}}MUCT, LFWcrop, \\ Celeb-A\end{tabular}      & Yes                             & Preserved                  \\ \hline
\end{tabular}
\caption{Summarizing the attribute suppression/anonymization algorithms in the literature.}
\label{Method_Comparison}
\end{table*}
%%%%%%%%%%%%%%%%%%%%%%%%%%%%%%%%%%%%%%%%%%%%%%%%%%%%%%%%%%%%%

Motivated by these observations, in this research, we raise an important question: ``Can we anonymize certain attributes for privacy preservation and \textit{fool} automated methods for attribute predictions?'' The answer to this question relates to $k$-anonymity literature where information of each individual cannot be differentiated from the $k-1$ individuals \cite{sweeney2002k}. It differs from the fact that in attribute anonymization, certain attributes are to be suppressed/anonymized while remaining ones are retained. For instance, if the images are uploaded to the driving license database to ascertain identity, it should not be used for any other facial analysis, except identity matching.     

%To address this challenge, in this research, we ask the question ``can we anonymize certain attributes from the face image for preserving privacy?". The research related to k-anonymity ensures that information of each individual cannot be differentiated from the k-1 individuals \cite{sweeney2002k}. However, in this research, the objective is to retain some attributes while anonymizing certain other information for every individual. For instance, if the images are uploaded to the driver's license database to ascertain identity, it should not be used for any other facial analysis. Figure \ref{fig:Introduction_Image} shows a person doing facial analysis or profiling from the face image used in ID card.

%%%%%%%%%%%%%%% Subection Related Work %%%%%%%%%%%%%%%%%%%%%%%%%%%%%

\subsection{Literature Review}

In the literature, privacy preservation in face images has been studied from two perspectives. Researchers have studied this problem in terms of privacy-preserving biometrics while others have termed it as attribute suppression. As mentioned above, face images reveal a lot of ancillary information for which the user may not have consented. In order to protect such ancillary information (soft biometrics), researchers have proposed several different methodologies. \cite{boyle2000effects} developed an algorithm to blur and pixelate the images in the video. \cite{newton2005preserving} have developed an algorithm for face de-identification in video surveillance such that the face recognition fails while preserving other facial details. \cite{gross2006model} have shown that distorting image via blurring and pixelation method gives poor results. To improve the results, they proposed model based face de-identification method for privacy protection.
%%%%%%%%%%%%%%%% Figure Literature Comparison %%%%%%%%%%%%%%%%%%
\begin{figure}
\centering
    \includegraphics[scale=0.36]{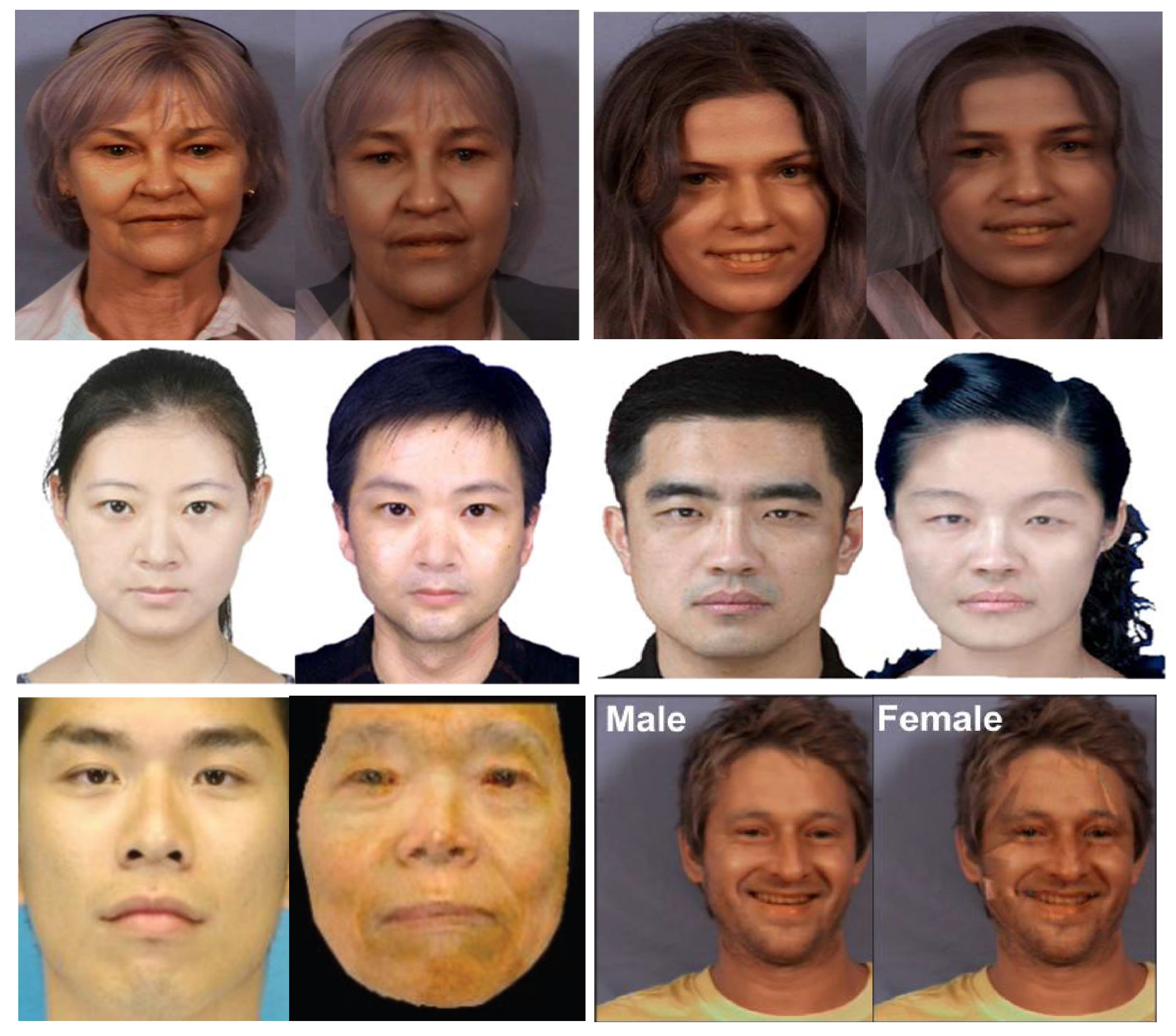}
	\caption{Demonstrating the attribute anonymized images generated by existing algorithms. First and third column images are original images while second and fourth column images are gender anonymized images \protect\cite{sim2015controllable}, \protect\cite{suo2011high}, \protect\cite{othman2014privacy}, \protect\cite{mirjalilisoft}.}
\label{fig:Literature}
\end{figure}
%%%%%%%%%%%%%%%%%%%%%%%%%%%%%%%%%%%%%%%%%%%%%%%%%%%%%%%%%%%%%%%
Some researchers have also worked on the privacy of soft biometrics. \cite{othman2014privacy} have proposed attribute privacy preserving technique, in which the soft biometrics attribute such as gender is ``flipped'' while preserving the identity for face recognition. In order to flip the gender, face morphing scheme is used in which the face of other opposite gender is morphed with the input image. As an extension of this work, \cite{mirjalilisoft} and \cite{mirjalili2017semi} have proposed Delaunay triangulation, and convolutional autoencoders based methods to flip gender information while preserving face identity. \cite{suo2011high} have presented an image fusion framework in which the template of opposite gender face image is taken for fusion with the candidate image while preserving face identity. \cite{jourabloo2015attribute} have developed an algorithm for de-identification of face image while preserving other attributes. For attribute preservation, it uses $k$ images (motivated by $k$-Same) which shares the same attributes for fusion. \cite{sim2015controllable} have proposed a method which independently controls the identity alteration and preserves the other facial attributes. It decomposes the facial attribute information into different subspaces to independently control these attributes. 
         
To anonymize the facial attributes, \cite{rozsa2017facial} have proposed deep learning based model for facial attribute prediction. A deep convolutional neural network is trained for each attribute separately and in order to test the robustness of the trained model, adversarial images are generated using fast flipping attribute (FFA) technique. As an extension of their work \cite{rozsa2016facial} have used FFA and  adversarial images are generated in which a facial attribute is flipped. They have observed that the few attributes are effected while flipping the targeted attribute. For instance, while changing the ‘wearing lipstick’ attribute, other attributes such as ‘attractive' and ‘heavy makeup’ are also flipped.

Based on the literature review, we observe that there are algorithms for single attribute anonymization. However, there are three major challenges in anonymizing multiple attributes. 
\begin{enumerate}
  \itemsep-0.3em
  \item While anonymizing facial attributes, there should be no visual difference between original and anonymized images.
  \item Selectively anonymizing few and retaining some attributes require a ``control'' mechanism. For example, gender and expression can be anonymized while retaining race and eye color and other attributes such as attractiveness and hair color may be in ``do not care'' condition.
  \item In applications involving matching faces for recognition, identity should be preserved while anonymizing attributes.
\end{enumerate}

The major limitation of these methods is that they do not address the first two challenges mentioned above. Existing algorithms primarily depend on a candidate image which is fused with the original image. This privacy preserving fusion leads to major transform and loss in visual appearance. \cite{rozsa2016facial} have addressed this issue but due to lack of any control mechanism, the other attributes are also suppressed, while suppressing one attribute.

%%%%%%%%%%%%% Subsection Research Contribution %%%%%%%%%%%%%%%%

\subsection{Research Contributions}
This research proposes a novel algorithm for privacy-preserving $k$-attribute anonymization using adversarial perturbation. The proposed algorithm jointly anonymizes and preserves multiple facial attributes without affecting the visual appearance of the image. The proposed algorithm is also extended for identity preserving $k$-attributes anonymization. Experiments on three databases and comparison with existing techniques showcase the efficacy of the proposed algorithm.
%%%%%%%%%%%%%%%%%%%%%%%% Figure Block Diagram %%%%%%%%%%%%%%%%%%%%%
\begin{figure}
\centering
    \includegraphics[scale=0.36]{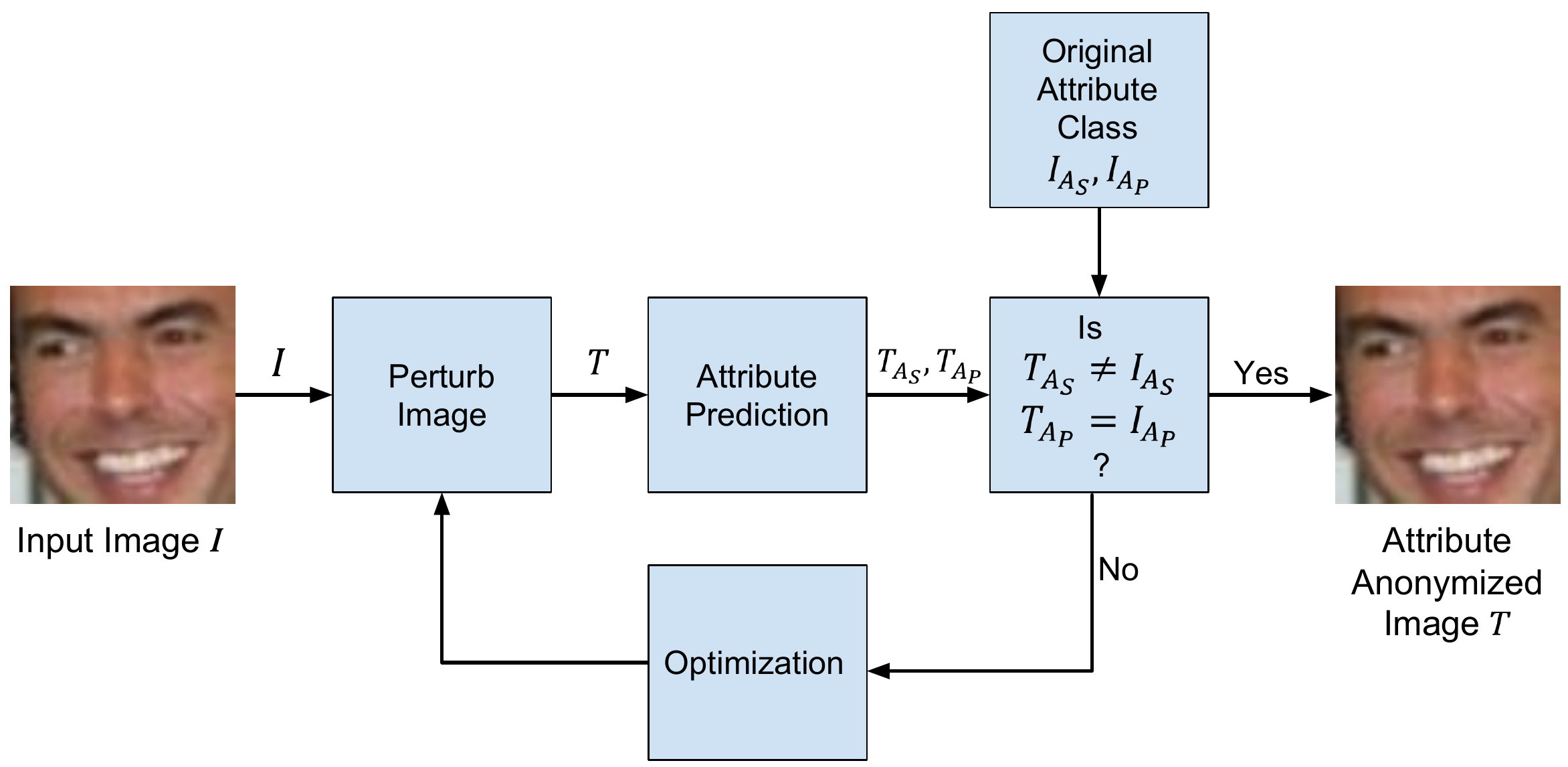}
	\caption{Illustrating the steps involved in the proposed algorithm. }
\label{fig:Block_Diagram}
\end{figure}
%First the original image will be given to the system followed by processing of it. Then the attribute classification will be done followed by comparison of original image attributes and processed image attributes. If the conditions are not satisfied then the image and the attribute score will be passed for optimization iteratively. Once the conditions are satisfied, we get the output anonymized image.
%%%%%%%%%%%%%%%%%%%%%%%%%%%%%%%%%%%%%%%%%%%%%%%%%%%%%%%%%%%%%%%%%%%

%%%%%%%%%%%%%%% Section Proposed Approach %%%%%%%%%%%%%%%%%%%%%%%%%
\section{Proposed Approach}
\label{Algo}
The problem statement can formally be defined as: ``create an image such that a set of pre-defined attributes are preserved while another set of pre-defined attributes are suppressed\rq\rq. As shown in Figure \ref{fig:Block_Diagram}, the proposed algorithm can change the prediction output of certain attributes from the true class to a different target class. The detailed description of the proposed algorithm is given below:

Let $\mathbf{I}$ be the original face image with $k$ number of attributes in the attribute set $\mathbf{A}$. For each attribute $\mathbf{A_i}$, there are $C_i$ number of classes. For instance, in attribute `Gender', two classes are \{Male, Female\} and `Expression' attribute has five classes, namely \{Happy, Sad, Smiling, Anger, Fearful\}. Mathematically, it is written as:

%\begin{equation}
%A_i = {A_i(C_i)}
%\end{equation}

\begin{equation}
\label{eq:1}
\mathbf{A} = \{\mathbf{A_1}(C_1), \mathbf{A_2}(C_2),...\; \mathbf{A_k}(C_k)\}
\end{equation}

Let $\mathbf{T}$ be the attribute anonymized image generated by adding perturbation $\mathbf{w}$ in the original image $\mathbf{I}$ where range of $\mathbf{T}$ is between $0$ to $1$. Mathematically, it is written as:
\begin{equation}
\mathbf{T} = \mathbf{I} + \mathbf{w}
\end{equation}
$$\text{such that} \quad \mathbf{T} \in [0,1]$$
To satisfy the above constraint, $tanh$ function is applied on $\mathbf{I} + \mathbf{w}$ as follows:
\begin{equation}
\label{eq:10}
\mathbf{T} = \frac{1}{2}(tanh(\mathbf{I} + \mathbf{w})+1)
\end{equation}

In an attribute suppressing/preserving application, let $\mathbf{A_S}$ and $\mathbf{A_P}$ be the sets of attributes to be suppressed and preserved where, $\mathbf{A} \geq (\mathbf{A_S} \cup \mathbf{A_P})$ and $\mathbf{A_S} \cap \mathbf{A_P} = \phi$. In $k-$ attribute anonymization task, it has to be ensured that the class of $\mathbf{A_S}$ attributes in $\mathbf{T}$ changes to some other class while preserving $\mathbf{A_P}$ attributes in image $\mathbf{T}$.

\begin{equation}
\label{eq:2}
\mathbf{T_{A_S}} \neq \mathbf{I_{A_S}}, \mathbf{T_{A_P}} = \mathbf{I_{A_P}}
\end{equation}

In order to suppress and preserve the sets of attributes $\mathbf{A_S}$ and $\mathbf{A_P}$ respectively, the distance between attributes of $\mathbf{A_P}$ in $\mathbf{I}$ and $\mathbf{T}$ is minimized, while the distance between attributes $\mathbf{A_S}$ is maximized. The objective function is thus represented as: 

\begin{equation}
\label{eq:3}
\text{min } \left[ D(\mathbf{I_{A_P}}, \mathbf{T_{A_P}}) - D(\mathbf{I_{A_S}}, \mathbf{T_{A_S}}) \right]
\end{equation}
$$\text{such that} \quad \mathbf{T_{A_S}} \neq \mathbf{I_{A_S}}, \mathbf{T_{A_P}} = \mathbf{I_{A_P}}$$
where, $D$ is the distance metric. To preserve the visual appearance of the image, the distance between $\mathbf{I}$ and $\mathbf{T}$ is also minimized. Experimentally, we found that the $\ell_2$ distance metric is most suitable for preserving the visual appearance of the image. Equation \ref{eq:3} is updated as:

\begin{equation}
\label{eq:4}
\text{min } \left \{ D(\mathbf{I_{A_P}}, \mathbf{T_{A_P}}) - D(\mathbf{I_{A_S}}, \mathbf{T_{A_S}}) + ||\mathbf{I} - \mathbf{T}||_2^2 \right \}
\end{equation}
$$\text{such that} \quad \mathbf{T_{A_S}} \neq \mathbf{I_{A_S}}, \mathbf{T_{A_P}} = \mathbf{I_{A_P}}$$

The first two constraints i.e. $\mathbf{T_{A_S}} \neq \mathbf{I_{A_S}}, \mathbf{T_{A_P}} = \mathbf{I_{A_P}}$ and the term $ \left[ D(\mathbf{I_{A_P}}, \mathbf{T_{A_P}}) - D(\mathbf{I_{A_S}}, \mathbf{T_{A_S}}) \right]$ in the above Equation is non linear. Therefore, to solve the same, an alternative function $f(\mathbf{T})$, inspired from \cite{carlini2017towards}, is used. The above Equation is thus written as:

\begin{equation}
\label{eq:5}
\text{min} \left \{ f(\mathbf{T}) + ||\mathbf{I} - \mathbf{T}||_2^2 \right \}
\end{equation}

Here, $f(\mathbf{T})$ attempts to preserve the attributes $\mathbf{A_P}$ and suppress attributes $\mathbf{A_S}$. 
There can be multiple cases for facial attribute anonymization, the objective function for each case is discussed as follows:

\paragraph{Case I - Single Attribute Anonymization:} This case formulates the scenario pertaining to changing the class of a single attribute i.e., the set $\mathbf{A_S}$ contains only one attribute, $\mathbf{U}$.
In order to change the class of a single attribute $\mathbf{U}$ with class $i$ to any other class $j$ where $(j \neq i)$, the objective function $f(\mathbf{T})$ is formulated as:

\begin{equation}
\label{eq:6}
f(\mathbf{T}) = max \left \{ 0, max(P(\mathbf{U}|\mathbf{T})) - P(U^j|\mathbf{T})\right \}
\end{equation}
where, $P(x)$ denotes a function giving the probability value of $x$. For our case, we have used the Softmax output score (discussed in Section 4). The term $max(P(\mathbf{U}|\mathbf{T}))$ used in Equation \ref{eq:6} outputs the maximum class score of attribute $\mathbf{U}$ and term $P(U^j|\mathbf{T})$ outputs the score of each class of attribute $\mathbf{U}$ except the $i^{th}$ class.
It is important to note that Equation \ref{eq:6} can also provide the target attribute class by giving a class label to the variable $j$.

%%%%%%%%%%%%%%%%%%%%%%%%%%%%%%%%%%%%%%%%%%%%%%%%%%%%%%%%%%%%%%%%%%%%
\paragraph{Case II - Multiple Attribute Preservation and Suppression:} This case formulates the scenario when the prediction output of multiple attributes, e.g. `Gender' and `Attractiveness' are suppressed while preserving other attributes e.g. `Ethnicity', `Wearing glasses', and `Heavy makeup'. In this specific example, the set $\mathbf{A_P}$ contains a list of three attributes whereas, $\mathbf{A_S}$ contains list of two attributes. For such scenarios of multiple attributes, the function $f(\mathbf{T})$ will be the summation of Equation \ref{eq:6} for each attribute. 
Mathematically, it is expressed as:
\begin{multline}
\label{eq:7}
f(\mathbf{T}) = \sum_{\mathbf{U}\in  \left \{\mathbf{A_S}, \mathbf{A_P} \right \}} max\{0, max(P(\mathbf{U}|\mathbf{T})) - \\ P(U^j|\mathbf{T})\}
\end{multline}

In order to preserve the attributes, the target attribute class will be same as input attribute class. The optimization method in Equation \ref{eq:7} can control the separation between target class score and next maximum class score by providing value `$-c$' in place of $0$ where $c \in [0, 1]$, i.e.

\begin{multline}
\label{eq:8}
f(\mathbf{T}) = \sum_{\mathbf{U}\in  \left \{\mathbf{A_S}, \mathbf{A_P} \right \}} max\{(-c, max(P(\mathbf{U}|\mathbf{T})) - \\ P(U^j|\mathbf{T}))\}
\end{multline}

\paragraph{Case III - Identity Preservation:} The third and the most relevant case with respect to face recognition applications, is preserving the identity $\mathbf{Id}$ of a person while suppressing an attribute. The objective function in this case can be written as:

\begin{equation}
\label{eq:9}
\text{min } \left \{f(\mathbf{T}) + ||\mathbf{I} - \mathbf{T}||_2^2 + \mathcal{D}(\mathbf{Id_I}, \mathbf{Id_{T}})\right \}
\end{equation} 

Here, $\mathbf{Id_I}$ and $\mathbf{Id_T}$ is the identity of the face images $\mathbf{I}$, and $\mathbf{T}$ respectively obtained from any face recognition algorithm, and $\mathcal{D}$ is the distance metric to match two face features.

%%%%%%%%%%%%%%% Experiment Details Table %%%%%%%%%%%%%%%%%
\begin{table}[]
\centering
\scriptsize
\begin{tabular}{|c|c|c|c|c|}
\hline
\multirow{2}{*}{\textbf{Experiment}}                             & \multirow{2}{*}{\textbf{Dataset}}                                    & \multirow{2}{*}{\textbf{\begin{tabular}[c]{@{}c@{}}\# Attributes \\ Anonymized\end{tabular}}} & \multicolumn{2}{c|}{\textbf{Attribute Anonymized}}                                                                                                   \\ \cline{4-5} 
                                                                 &                                                                      &                                                                                        & \textbf{Suppressed}                                                       & \textbf{Preserved}                                                       \\ \hline
\begin{tabular}[c]{@{}c@{}}Single \\ Attribute\end{tabular}      & \begin{tabular}[c]{@{}c@{}}MUCT, \\ Celeb-A, \\ LFWcrop\end{tabular} & 1                                                                                      & Gender                                                                    & -                                                                        \\ \hline
\begin{tabular}[c]{@{}c@{}}Multiple \\ Attributes\end{tabular}   & Celeb-A                                                              & 3, 5                                                                                   & \begin{tabular}[c]{@{}c@{}}Gender, \\ Attractive, \\ Smiling\end{tabular} & \begin{tabular}[c]{@{}c@{}}Heavy makeup, \\ High cheekbones\end{tabular} \\ \hline
\begin{tabular}[c]{@{}c@{}}Identity \\ Preservation\end{tabular} & \begin{tabular}[c]{@{}c@{}}MUCT, \\ LFWcrop\end{tabular}             & 1+1                                                                                    & Gender                                                                    & Identity                                                                 \\ \hline
\end{tabular}
\caption{The experiments are performed pertaining to three cases to showcase the effectiveness of the proposed algorithm.}
\label{Experimental_details}
\end{table}
%%%%%%%%%%%%%%

\begin{figure}
\centering
    \includegraphics[scale=0.7]{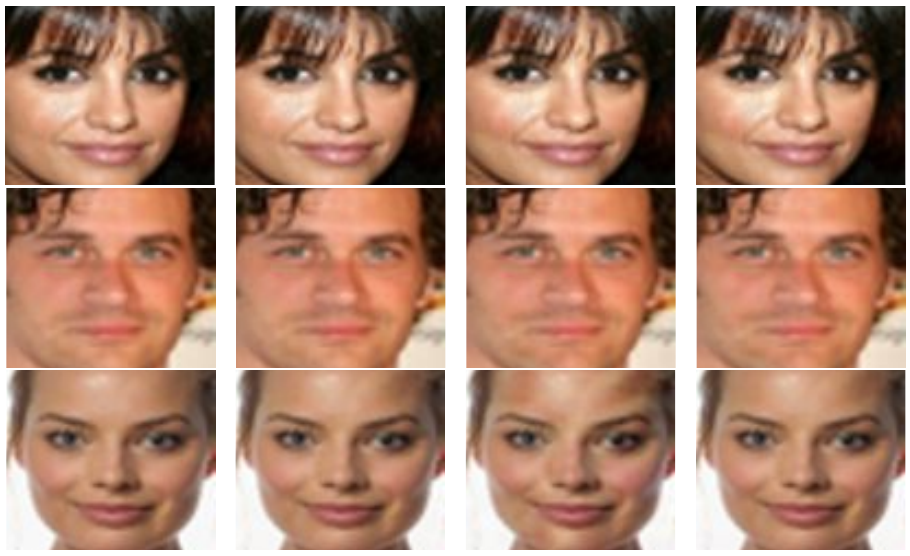}
	\caption{This figure shows the images before and after anonymizing attributes of Celeb-A dataset. First column represents the original images while second, third, and fourth columns show one, three, and five attributes anonymized images.}
\label{fig:CelebA_Face_Collage}
\end{figure}

The proposed formulation of $k$-anonymizing attributes can be viewed as adding adversarial noise such that some attributes are suppressed while preserving selected attributes and identity information. Figure \ref{fig:CelebA_Face_Collage}, shows some examples of attribute anonymization using the proposed algorithm.
%%%%%%%%%%%%%%%%%%%%%%%% Section Dataset %%%%%%%%%%%%%%%%%%%%%%%%%%%
\section{Datasets and Experimental Details}

%%%%%%%%%%%%%%%%%% CONFUSION MATRIX %%%%%%%%%%%%%%%%%%%% CM
\begin{table*}[]
\centering
\scriptsize

\begin{tabular}{|c|c||c|c|c||c|c|c||c|c|c|}
\hline                                                                  & \textbf{}                                                                       & \textbf{\begin{tabular}[c]{@{}c@{}}Attribute \\ Class\end{tabular}} & \multicolumn{2}{c||}{\textbf{Prediction}} & \textbf{\begin{tabular}[c]{@{}c@{}}Attribute \\ Class\end{tabular}} & \multicolumn{2}{c||}{\textbf{Prediction}}                         & \textbf{\begin{tabular}[c]{@{}c@{}}Attribute \\ Class\end{tabular}} & \multicolumn{2}{c|}{\textbf{Prediction}}                               \\ \hline
\multirow{5}{*}{\begin{tabular}[c]{@{}c@{}}Ground\\ Truth\end{tabular}} &                                                                                 &                                                                     & Male               & Not Male              &                                                                     & Smiling & \begin{tabular}[c]{@{}c@{}}Not Smiling\end{tabular} &                                                                     & Attractive & \begin{tabular}[c]{@{}c@{}}Not Attractive\end{tabular} \\ \cline{2-11} 
                                                                        & \multirow{2}{*}{\begin{tabular}[c]{@{}c@{}}Before\\ Anonymization\end{tabular}} & Male                                                                & 87.70               & 12.30                 & Smiling                                                             & 64.59    & 35.41                                                   & Attractive                                                          & 89.31       & 10.69                                                      \\ \cline{3-11} 
                                                                        &                                                                                 & Not Male                                                              & 19.64               & 80.36                & \begin{tabular}[c]{@{}c@{}}Not Smiling\end{tabular}              & 24.66    & 75.34                                                   & \begin{tabular}[c]{@{}c@{}}Not Attractive\end{tabular}           & 28.41       & 71.59                                                      \\ \cline{2-11} 
                                                                        & \multirow{2}{*}{\begin{tabular}[c]{@{}c@{}}After \\ Anonymization\end{tabular}} & Male                                                                & 3.89                  & 96.11                & Smiling                                                             & 0.02       & 99.98                                                   & Attractive                                                          & 0.28         & 99.72                                                      \\ \cline{3-11} 
                                                                        &                                                                                 & Not Male                                                              & 100              & 0                   & \begin{tabular}[c]{@{}c@{}}Not Smiling\end{tabular}              & 99.90    & 0.10                                                     & \begin{tabular}[c]{@{}c@{}}Not Attractive\end{tabular}           & 99.59      & 0.41                                                        \\ \hline
\end{tabular}
\caption{Confusion matrix displaying classification accuracies (\%) of before and after suppression of three attributes together on the CelebA dataset}
\label{Confusion_Mat_3_Attrib}
\end{table*}

The proposed algorithm is evaluated on three datasets: MUCT \cite{Milborrow10}, LFWcrop \cite{LFWTech}, and CelabA\cite{liu2015faceattributes}. Comparison has been performed with the algorithm proposed by \cite{mirjalilisoft}. The details of the databases are described as follows:

\textbf{MUCT dataset} contains 3,755 images of 276 subjects out of which 131 are male and 146 are female captured under varying illuminations using five webcams. Viola-Jones \cite{viola2005detecting} face detector is applied to all images and the detector fails to detect 49 face images. Therefore, only 3,706 images are used for further processing. \textbf{LFWcrop dataset} contains 13,233 face images of 5,749 subjects. In order to evaluate the performance, view 2 of the dataset which consist of 6,000 pairs of images, has been considered. Out of these 6,000 pairs of images, only one image from each pair has been selected for anonymization. \textbf{CelebA dataset} contains 202,599 face images of celebrities. From this dataset, only the test set of 19,962 images has been considered for attribute anonymization. The results of anonymizing a single attribute are shown on all three databases while the results of anonymizing multiple attributes are shown on CelebA dataset.

As shown in Table \ref{Experimental_details}, three experiments are performed, one corresponding to each case discussed in Section \ref{Algo}. For attribute anonymization task, white box attacks are performed whereas for identity preservation, black box attack is performed. The attributes are anonymized corresponding to the attribute classification model while the identities are preserved according to face recognition algorithm. For attribute classification, we have selected fine-tuned VGGFace \cite{parkhi2015deep} and for identity preservation, OpenFace \cite{amos2016openface} model is used. The distance metrics used with both the models is Euclidean distance. The performance of face recognition is evaluated on both OpenFace \cite{amos2016openface} and VGGFace (black box model) \cite{parkhi2015deep}. 

%As shown in Table \ref{Experimental_details}, three experiments are performed, one for each case, as discussed in Section \ref{Algo}. For attribute anonymization task, white box attacks are performed whereas for identity preservation, black box attack is performed. Attributes are anonymized corresponding to the attribute classification model i.e. fine tuned VGGFace \cite{parkhi2015deep} whereas identities are preserved corresponding to OpenFace \cite{amos2016openface} model and the performance is evaluated on both OpenFace  \cite{amos2016openface} and VGGFace (black box model)  \cite{parkhi2015deep}. The distance metrics used in this research is Euclidean distance.
\paragraph{Implementation details:} The proposed algorithm is implemented in Tensorflow with 1080 Ti GPU. For learning the perturbation, L2 attack has been performed with Adam optimizer. The learning rate is set 0.01 and number of iterations used is 10000. 

%The protocol for each dataset and attribute classification model is discussed In Appendix A.

%%%CelebA 5 Attrib
\begin{figure*}
\centering
\small
    \includegraphics[scale=0.23]{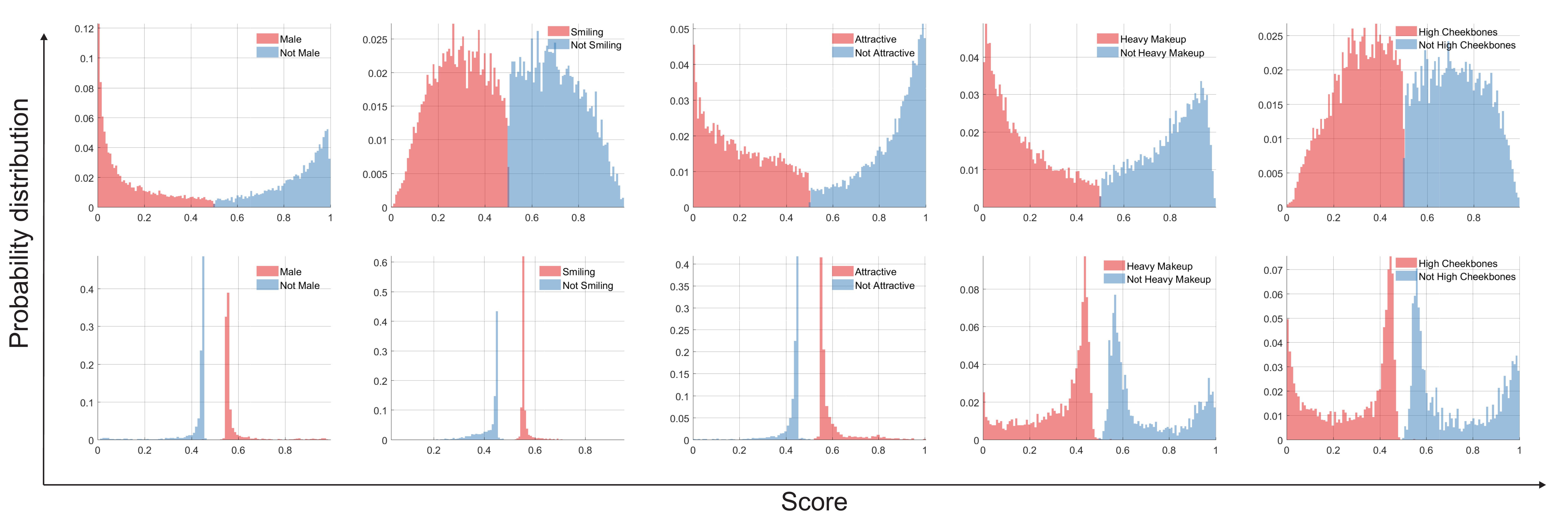}
	\caption{Comparing the attribute class score distributions of images before and after anonymization of 5 attributes together on the CelebA dataset. The first row distributions pertain to the original images whereas, second row distributions corresponds to the anonymized images.}
\label{fig:CelebA_5_Attrib}
\end{figure*} 
%%%%%%%%%%%%%%%%%%%%%%%%%%%%%%%%%%%%%%%%%%
 %%%%%%% CelebA 1 Attrib
\begin{figure}
\centering
\small
    \includegraphics[scale=0.28]{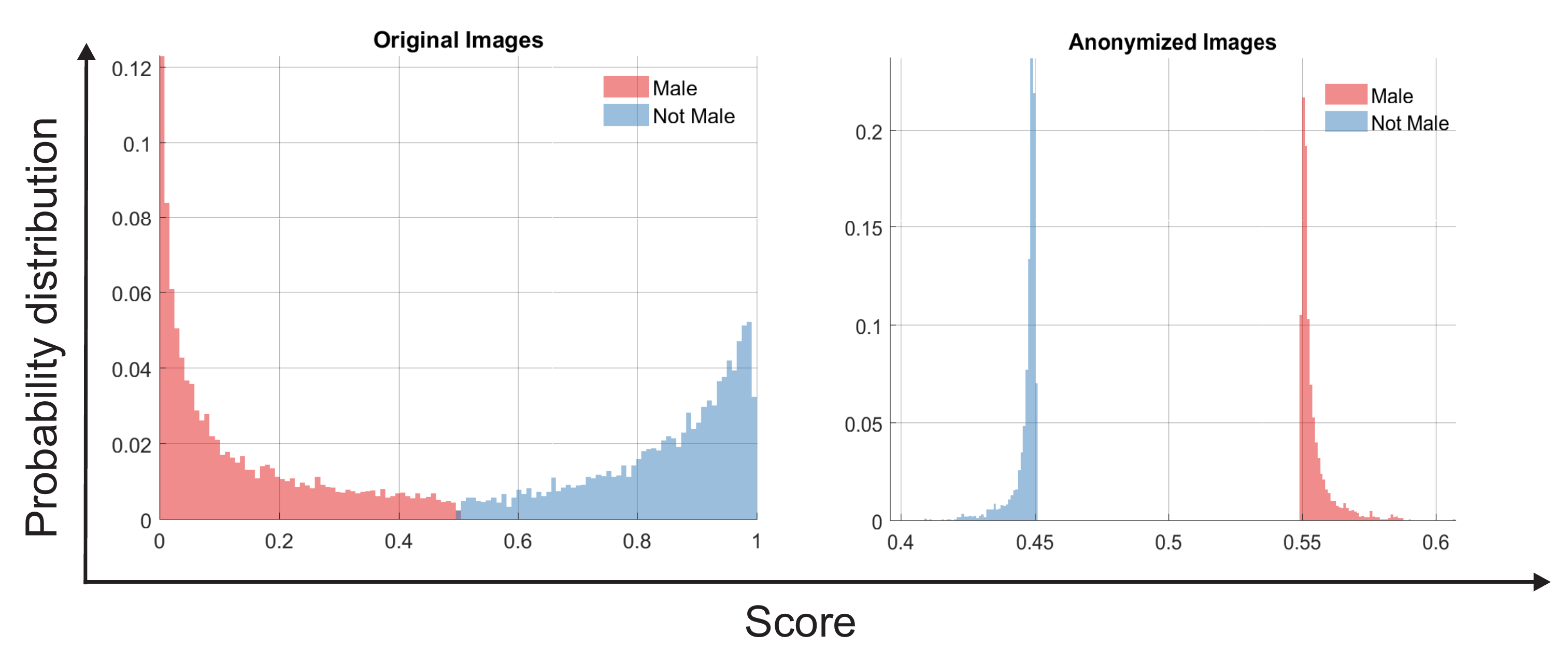}
	\caption{Comparing score distributions before and after anonymizing 'Gender' attribute on the CelebA dataset. }
\label{fig:CelebA_1_Attrib}
\end{figure}
%%%%%%%%%%%%%%%%%%%%%%%%%%  Result Section %%%%%%%%%%%%%%%
\section{Performance Evaluation}
The proposed model is evaluated for single and multiple attribute anonymization. For multiple attribute anonymization, two types of experiments are performed: attribute suppression, and attribute suppression $+$ preservation. In case of single attribute anonymization, the performance of the proposed algorithm has been analyzed for the tasks of attribute suppression as well as identity preservation. To generate the attribute anonymized images, only those samples have been considered which are correctly classified by the attribute classification model. 

%%%%%%%%%%%%%% Subsection Performance for Single Attribute %%%%%%%%%%%%%%%%
\subsection{Multiple Attribute Anonymization}
The proposed algorithm for multiple attribute anonymization is evaluated on the CelebA dataset. Three attributes are considered for the task of attribute suppression, while attribute suppression and preservation is performed with five attributes.

\paragraph{Attribute Suppression:} Experiments have been performed with three attributes i.e. `Gender', `Smiling', `Attractive'. The confusion matrix in Table \ref{Confusion_Mat_3_Attrib} shows that in over 96\% of the images, attributes are correctly suppressed. After applying the proposed algorithm for the attribute `Attractive', the classification accuracy dropped from 89.31\% to 0.28\%; thereby showcasing the efficacy of the proposed technique for the task of attribute suppression. 

\paragraph{Attribute Suppression and Preservation:} The results for five attributes `Gender', `Attractive', `Smiling', `Heavy Makeup' and `High Cheekbones' are shown in Figure \ref{fig:CelebA_5_Attrib}. Three attributes are suppressed i.e. `Gender', `Attractive' and `Smiling', while remaining two are preserved. The confidence value (\textit{`c'} value as defined in Equation \ref{eq:8}) is set to 0.1. In Figure \ref{fig:CelebA_5_Attrib}, the first and second row histograms show the attribute class score distributions before and after anonymization of images. It is observed that the attribute class score distributions of first three attributes `Gender', `Attractive' and `Smiling' are flipped while the class score distributions of `Heavy Makeup' and `High Cheekbones' are preserved. This illustrates the utility of the proposed algorithm for multiple attributes suppression and preservation. Figure \ref{fig:CelebA_Face_Collage} presents the original and anonymized images for one, three, and five attributes. The similarity of visual appearance between original and modified images further strengthens the usage of the proposed algorithm. 
%%% Add bar graph %%%%

% of 'Gender' attribute only. The 3 attribute anonymization is performed for suppression of attributes only while 5 attributes anonymization is performed for both suppression and preservation of attributes. 

%%%%%%%%%%%%%%%%%%%%%%%%%%%%

%%%%%

%%%%%%%%%%%%%%%%%%%%%%%%%%%%%%%%%%%%%%%
\subsection{Attribute Suppression with Identity Preservation}
To evaluate the performance of single attribute suppression with identity preservation, experiments are performed on MUCT, CelebA and LFWcrop datasets. The results of single attribute suppression and attribute suppression along with identity preservation are discussed below.

\begin{figure}
\centering
    \includegraphics[scale=0.65]{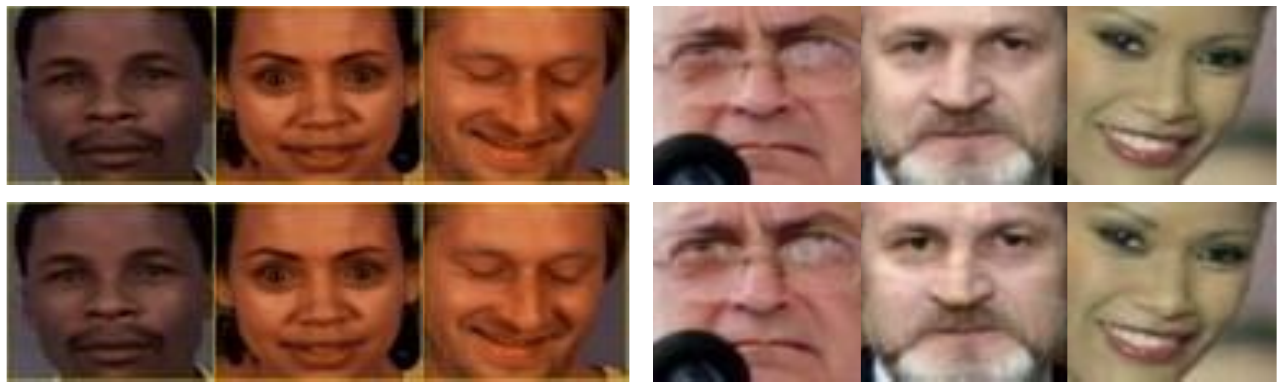}
	\caption{Examples of before and after anonymization of `Gender' attribute on MUCT and LFWcrop datasets. First row images are original images, second row images are `gender' anonymized images.}
\label{fig:MUCT_LFW_Face_Collage}
\end{figure}

\paragraph{Single Attribute Suppression:} Figure \ref{fig:CelebA_1_Attrib} shows the result of the `Gender' attribute suppression on CelebA dataset. It can be observed that the score distribution of attributes are completely flipped before and after anonymization. 
The confusion matrix for MUCT dataset before and after anonymization is shown in Table \ref{Confusion_Compl}. On comparing the results with the algorithm proposed by \cite{mirjalilisoft}, it can be observed that after anonymization, all samples are misclassified by the proposed method. It is important to note that the method proposed by \cite{mirjalilisoft} is dependent on the fusion of the other candidate images which embeds visual distortion in the image. On the other hand, in the proposed method, the anonymization of attribute is independent of another candidate image, thus resulting in limited visual distortions. Some example images of a suppressed attribute on MUCT and LFWcrop datasets are demonstrated in Figure \ref{fig:MUCT_LFW_Face_Collage}. These samples show that there are minimal effect on visual appearance of the image after attribute anonymization.

\begin{figure}[!t]
\centering
    \includegraphics[scale=0.38]{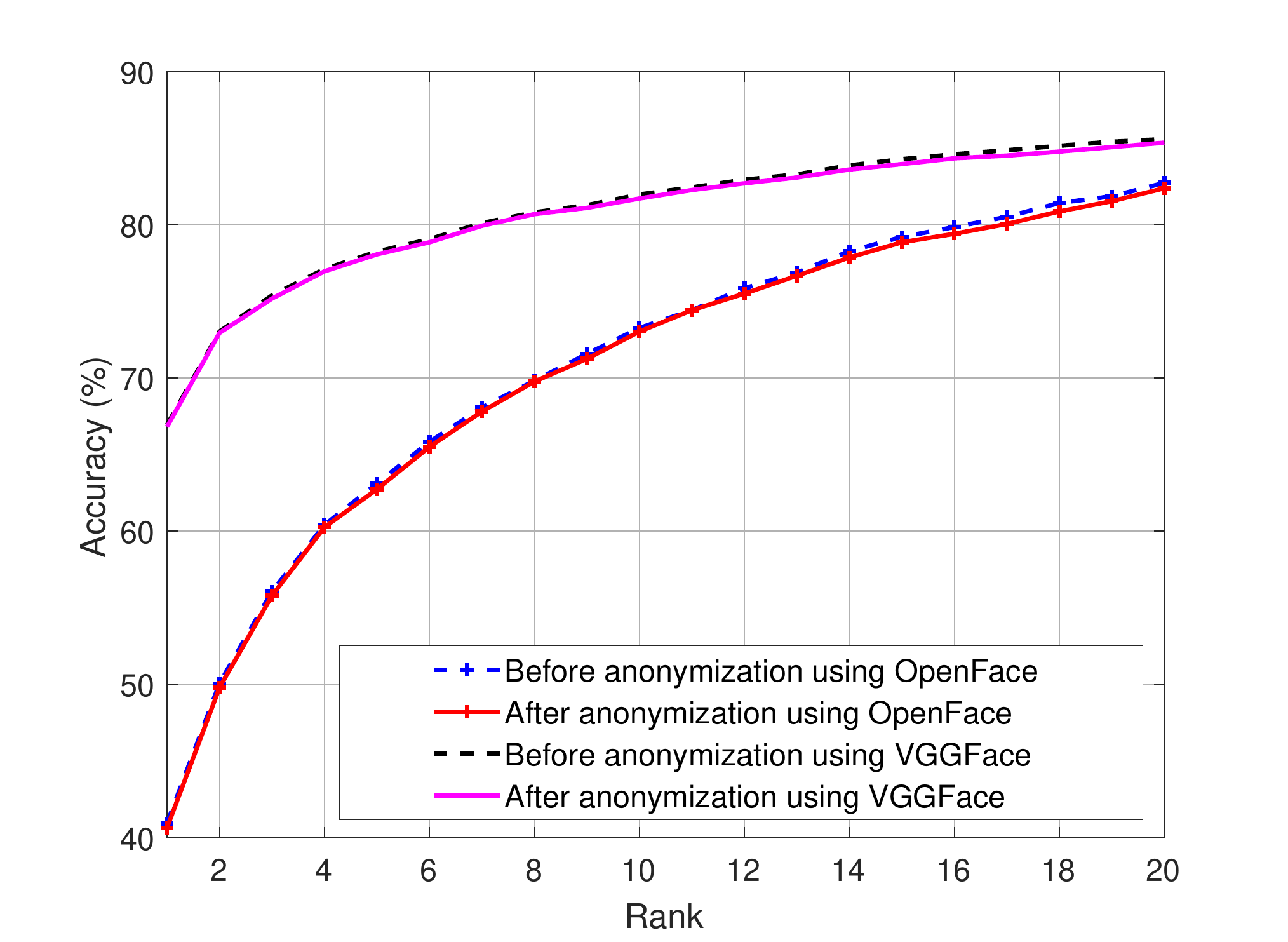}
	\caption{CMC curve showing original to original face matching vs original to anonymized face matching on the MUCT dataset.}
\label{fig:CMC_MUCT}
% \end{figure}
% \begin{figure}
% \centering
    \includegraphics[scale=0.38]{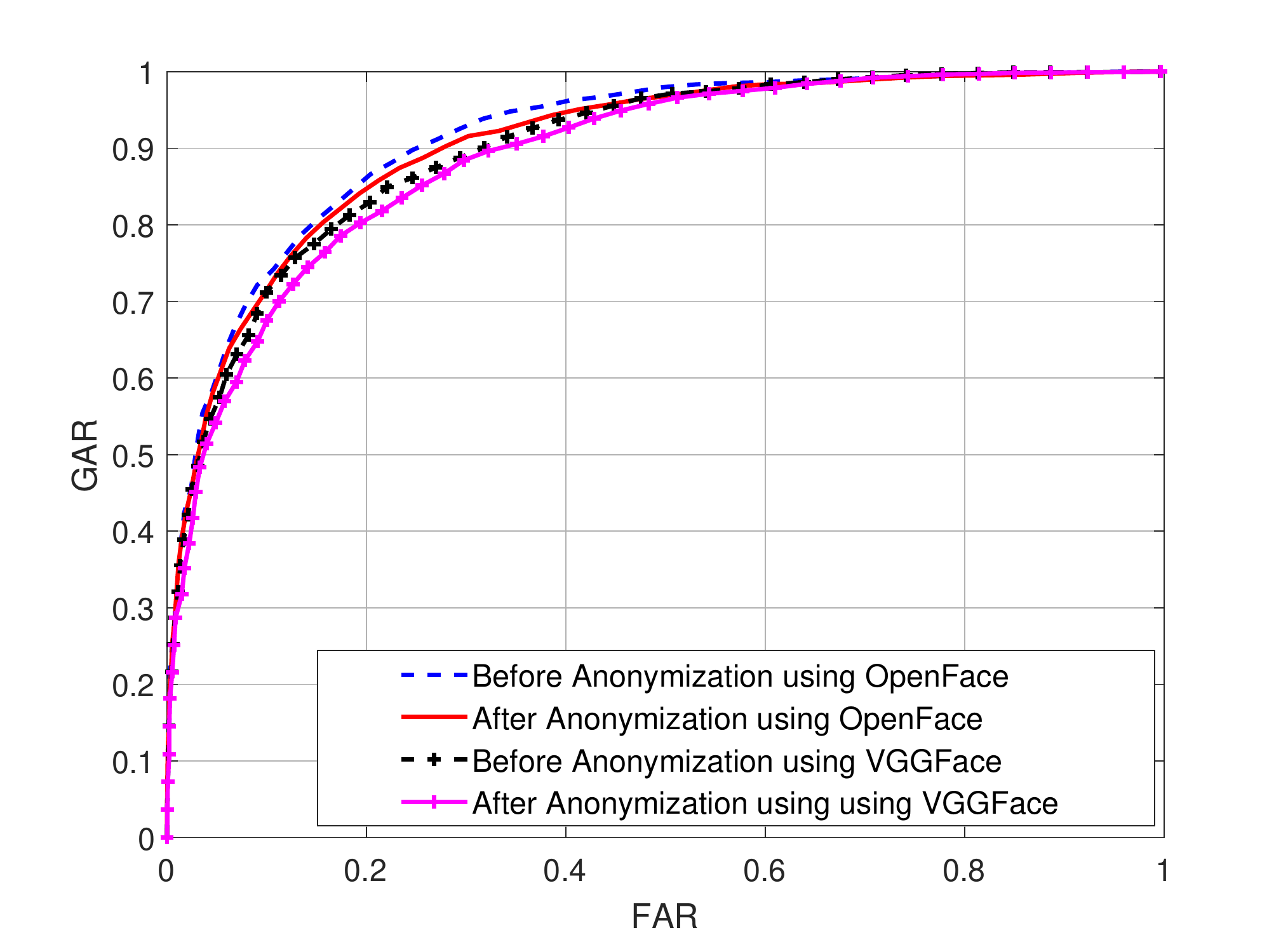}
	\caption{ROC curve showing face recognition performance on the LFWcrop Dataset}
\label{fig:LFW_6000_ROC}
\end{figure}

\paragraph{Identity Preservation:} To evaluate identity preservation performance, while anonymizing an attribute, experiments are performed on both MUCT and LFWcrop datasets. For MUCT dataset, a single image gallery and two probe sets are used. The first probe set consist of original images while the second probe set contains the corresponding `Gender' suppressed images of probe set 1. Two widely use face recognition models i.e. OpenFace \cite{amos2016openface} and VGGFace \cite{parkhi2015deep} are used for (i) original to original face matching (i.e., gallery to probe set 1) and (ii) original to anonymized face matching (i.e., gallery to probe set 2). The Cumulative Match Characteristic (CMC) curves summarizing the performance of face recognition before and after suppressing the attributes are show in Figure \ref{fig:CMC_MUCT}. The low variation observed in accuracy for recognition before and after anonymization motivates the applicability of the proposed algorithm for identity preservation as well.

% \usepackage{multirow}
% \usepackage{multirow}
\begin{table}[]
\centering
\scriptsize
\begin{tabular}{|c|c|p {12 mm}|c|c|c|}
\hline
\textbf{Anonymization}                                                                   & \textbf{\begin{tabular}[c]{@{}c@{}}Attribute\\ Class\end{tabular}} & \multicolumn{2}{c|}{\textbf{\begin{tabular}[c]{@{}c@{}}\cite{mirjalilisoft}\end{tabular}}} & \multicolumn{2}{c|}{\textbf{Proposed}} \\ \hline
                                                                                         &                                                                    &\centering  Male              & Female                                        & Male              & Female             \\ \hline
\multirow{2}{*}{\textbf{\begin{tabular}[c]{@{}c@{}}Before\end{tabular}}} & Male                                                               & \centering 1762                 & 17                                           & 1741                 & 87               \\ \cline{2-6} 
                                                                                         & Female                                                             &\centering  521             & 1300                                          & 252             & 1626                  \\ \hline
\multirow{2}{*}{\textbf{\begin{tabular}[c]{@{}c@{}}After\end{tabular}}} & Male                                                               & \centering 276              & 1503                                          & 0                & 1828               \\ \cline{2-6} 
                                                                                         & Female                                                            & \centering 1255             & 566                                          & 1878             & 0                 \\ \hline
\end{tabular}
\caption{Comparison of confusion matrix of the proposed algorithm with \protect\cite{mirjalilisoft} on the MUCT dataset.}
\label{Confusion_Compl}
\end{table}

%% Figure MUCT

%%%%%%%%%%%%%%%%%%%%%%%%%%%%%%%%%%%%%%%%%%

%%%%%%%%%%%%%%%%%%%%% Subsection Performance Identity Preservation %%%%%%%%%%%%%%%

%%%%%%%%%%%%%%%%%%%%%% CMC %%%%%%%%%%%%%%%%%%%%%%%%%%%%
%%%%%%%%%%%%%%%%%%%%%%%%%%%%%%%%%%%%%%%%%%%%%%%%%%%%%%%%%%%%%%

Figure \ref{fig:LFW_6000_ROC} shows the Receiver Operating Characteristic (ROC) curves for face verification on LFWcrop dataset. Experiments are performed on (i) 6000 original pair images, and (ii) 6000 original-anonymized pairs. Face verification is performed using OpenFace \cite{amos2016openface} and VGGFace model \cite{parkhi2015deep}. The results show minimal effect on the face verification performance before and after attribute anonymization.
%%%%%%%%%%%%%%%%%%%%%  ROC  %%%%%%%%%%%%%%%%%%%%%%%%%%%%%%%%%%%

%%%%%%%%%%%%%%%%%% Conclusion %%%%%%%%%%%%%%%%%%%
\section{Conclusion}
Attribute anonymization while preserving identity has several privacy preserving applications. This paper presents a novel algorithm based on adversarial noise addition concept such that selected attributes (or features) are anonymized and selected attributes (including identity information) are preserved for automated processing. Experiments are performed on CelebA, LFWcrop and MUCT databases with three different application scenarios. The results demonstrate that the proposed algorithm can handle multiple attribute anonymization process without affecting visual appearance and face recognition performance.  

\section*{Acknowledgments}
Vatsa and Singh are partially supported through Infosys Center for AI at IIIT-Delhi.
\bibliographystyle{named}
\bibliography{ijcai18}

\end{document}